# Exploring System Performance of Continual Learning for Mobile and Embedded Sensing Applications

Young D. Kwon
University of Cambridge
ydk21@cam.ac.uk

Jagmohan Chauhan
University of Southampton
J.Chauhan@soton.ac.uk

Abhishek Kumar
University of Helsinki
abhishek.kumar@helsinki.fi

Pan Hui
HKUST
University of Helsinki
panhui@cse.ust.hk

Cecilia Mascolo
University of Cambridge
cm542@cam.ac.uk

## ABSTRACT

Continual learning approaches help deep neural network models adapt and learn incrementally by trying to solve catastrophic forgetting. However, whether these existing approaches, applied traditionally to image-based tasks, work with the same efficacy to the sequential time series data generated by mobile or embedded sensing systems remains an unanswered question.

To address this void, we conduct the first comprehensive empirical study that quantifies the performance of three predominant continual learning schemes (i.e., regularization, replay, and replay with examples) on six datasets from three mobile and embedded sensing applications in a range of scenarios having different learning complexities. More specifically, we implement an end-to-end continual learning framework on edge devices. Then we investigate the generalizability, trade-offs between performance, storage, computational costs, and memory footprint of different continual learning methods.

Our findings suggest that replay with exemplars-based schemes such as iCaRL has the best performance trade-offs, even in complex scenarios, at the expense of some storage space (few MBs) for training examples (1% to 5%). We also demonstrate for the first time that it is feasible and practical to run continual learning on-device with a limited memory budget. In particular, the latency on two types of mobile and embedded devices suggests that both incremental learning time (few seconds - 4 minutes) and training time (1 - 75 minutes) across datasets are acceptable, as training could happen on the device when the embedded device is charging thereby ensuring complete data privacy. Finally, we present some guidelines for practitioners who want to apply a continual learning paradigm for mobile sensing tasks.

## 1 INTRODUCTION

Deep learning has revolutionized the performance of various disciplines, including mobile and embedded systems applications. This is particularly true for applications relying on continuous streams of sensor data such as activity recognition [13], mental health, and wellbeing [30], gesture recognition [10], tracking and localization [19]. However, a crucial characteristic common to the above applications is the need for a trained model to adapt to accommodate new classes and to a dynamically changing environment. In these settings, the ability to *continually* learn [20, 35], that is, to learn consecutive tasks without forgetting how to perform previously learned tasks, becomes essential. Let us consider an example. Alice has a deep learning model deployed on her smartphone for human activity recognition (HAR) to recognize simple activities such as sitting and standing. As time passes, the model might want to learn new activities such as walking to be more beneficial to a very active Alice. A static model will learn new activities but will fail to predict older activities correctly due to *catastrophic forgetting* (CF) [32]. CF means the abrupt and near-complete loss of knowledge obtained from previous tasks when the model learns new tasks. Specifically, weights in a model important to previous task A (i.e., previous task) are changed to optimize towards task B (i.e., new task), which often leads to the degradation of task A's performance. With addressing CF issues, continual learning [35] allows deep learning models to learn incrementally (adapt or accommodate new classes/behaviors) and obviates the need to be trained every time from scratch, which might waste valuable resources on Alice's device.

In practice, enabling deep learning models to continually learning is very challenging due to the CF problem. Since CF was first identified in Multi-Layer Perceptrons (MLPs), many researchers have proposed methods to mitigate it [22, 28, 29, 40, 54] and evaluate it using small and large datasets [20, 36, 42]. However, the proposed methods are mainly evaluated in the field of computer vision with MLPs or Convolutional Neural Networks (CNN) based deep learning models. *It is unclear whether these methods are viable in sensor-based applications, where the modality of the data is significantly different from images, and sequence information needs to be captured [38].* Moreover, *most of the existing Incremental Learning (IL)[1] techniques [26] do not take into account the resource requirements of these devices*, which may make them inapplicable to embedded and mobile systems deployments. There is a clear need to understand the resource consumption limitations of existing continual learning methods to see if they are applicable to resource-constrained edge platforms.

To address the aforementioned limitations of prior work, we conduct the first systematic study to investigate the CF problem on mobile and embedded sensing applications using various IL methods. **First,** we employ three datasets from the widely researched application of Human Activity Recognition (HAR) [5] based on accelerometer, gyroscope, and magnetometer data. Next, we include two datasets from Gesture Recognition (GR) [56] based on surface electromyography (sEMG). We further incorporate an Emotion Recognition (ER) dataset [39] based on speech among audio sensing tasks to make our results generalizable to different modalities across diverse applications. **Second,** we examine trade-offs of studied IL methods in terms of their performance, storage footprint, computational costs, and the peak memory limit to consider the feasibility and applicability of the IL methods on mobile and embedded devices. To investigate the system limitations imposed by different configurations of IL, we implemented the IL framework on two types of devices with different specifications – an Nvidia Jetson Nano GPU (used in mobile robotics and tablets) and a smartphone (One Plus 7 Pro) CPU – with respect to computational costs, storage, and memory footprint.

Overall, the major contributions and findings of this paper are:

**First,** we conduct a systematic investigation of the CF problem on mobile and embedded applications using six state-of-the-art IL methods falling under three paradigms: **regularization** ((1) Elastic Weight Consolidation: EWC [22], (2) Synaptic Intelligence: SI [55], and (3) Online EWC [43]), **replay** ((4) Learning without Forgetting: LwF [28]), and **replay with exemplars** ((5) Incremental Classifier and Representation Learning: iCaRL [40] and (6) Gradient Episodic Memory: GEM [29]). In addition, to make our study generalizable across different modalities of data, we perform analysis on six datasets of three different sensing applications (HAR, GR, and ER).

**Second,** to evaluate CF in real-life scenarios, we employ Sequential Learning Tasks (SLTs), successively learning two or more subtasks $D_1, ..., D_k$, instead of learning a single task $D$ [36]. Learning new tasks continuously becomes vital since the number of classes (activities or users) and the environments of edge applications often change over time. We adopt a class-incremental learning setup where each task contains distinct classes, which fits well with practical application scenarios (see §3.1 for detail). Specifically, we try three scenarios: adding only one class to a base classifier (simple), adding half of the classes, $N/2$, to a base classifier at once (mildly complex), and a very practical (complex) scenario where half of the classes, $N/2$, are added incrementally to a base classifier one by one, where $N$ is the total number of classes. Through extensive experiments, we find that all IL methods perform well when presented with simple scenarios but fail in the complex scenario, except for iCaRL. The main reason for iCaRL's strong performance is its use of exemplar samples. To the best of our knowledge, we are the first to train and implement IL methods to run on mobile and embedded systems, with the aim to build an end-to-end on-device continual learning system and to evaluate trade-offs of studied IL methods in terms of their performance, storage, and computational costs, as well as the peak memory usage.

**Third,** we find that iCaRL and GEM require a modest amount of storage, which seemingly is not an issue on many modern devices as they support a large amount of storage (in order of a few GBs). Even at a maximum number of stored exemplars (i.e., 20% - 40% of training samples), iCaRL and GEM require only 2 MB–115 MB. However, GEM and EWC-based algorithms are computationally expensive in that the average IL time varies from 46.3–2,660 seconds on Jetson Nano. For all other algorithms, it ranged from 8.46–150 seconds on both Jetson Nano and a smartphone. iCaRL, in particular, needs less than a minute on a smartphone to do IL on a per-task basis and operates within a reasonable peak memory overhead (196–2,127 MB). In sum, our study shows that simple deep learning architectures such as one and two-layer long short-term memory (LSTM) [17] can be trained entirely on the smartphone, thereby ensuring complete user privacy.

**Finally,** based on our findings, we present a series of lessons and guidelines to help practitioners and researchers in their use of continual deep learning for mobile sensing applications.

In addition to the above contributions, we adapt the experimental protocol proposed in [36] which considers learning only two tasks. We extend this protocol so that it can incorporate any number of tasks $(D_1, ..., D_k)$ in an incremental manner and identify the best performing IL model by permutating a set of hyper-parameters and IL-method-specific parameters (see §3.3 for detail). Finally, we believe that our work and findings open the door to the use of continual learning in edge devices and applications.

## 2 RELATED WORK

We begin by reviewing continual learning approaches and empirical studies to evaluate them, followed by applications of deep learning in the mobile and edge sensing domain.

### 2.1 Continual Learning

Continual learning studies the ability to learn over time from a coming stream of data by incorporating new knowledge while retaining previously learned experiences [35]. Continual learning is also called incremental learning (IL) [40], lifelong learning [35], and sequential learning [32]. In a continual learning setup, learning methods typically suffer from CF [31, 32], that is, a learned model experiences performance degradation on previously learned task(s) (e.g., task A) as information relevant to a new task (e.g., task B) is incorporated. It is because the learned parameters of the network that are optimized to perform well in task A (i.e., important weights to task A) are changed to maximize/minimize the objective/loss of task B. In recent years, many researchers have focused on solving the CF issue by proposing a range of IL approaches. The first group of approaches is a *regularization-based* method [22, 43, 45, 55] where regularization terms are added to the loss function to minimize changes to important weights of a model for previous tasks to prevent forgetting. Another group of approaches is a *replay-based* method [28] where model parameters are updated for learning a representation by using training data of the currently available classes, which is different from *replay with exemplars-based* method [23, 29, 40] where updating the model requires training data from the new class and also few training samples from earlier classes.

---
[1]In this work, we use continual learning (CL) and incremental learning (IL) interchangeably.

The proposed IL methods to solve CF are empirically evaluated using small and large datasets [20, 36]. However, these empirical studies either adopt only a few methods [20, 36] or neglect resource constraints of mobile and embedded devices with respect to storage and latency [18, 36]. To fill this gap, we perform a systematic study on six most cited (or state-of-the-art) IL methods from three representative categories of IL approaches with three continual learning scenarios with different difficulties. Also, we conduct the first comprehensive study of generalizability and trade-offs between performance, storage, and computational costs among the studied IL methods on mobile and embedded devices.

## 2.2 Deep Learning for Mobile Sensing Systems

Deep learning is increasingly being applied in mobile and embedded systems as it achieves state-of-the-art performances on many sensing applications such as activity recognition [14, 33], gesture recognition [37], and audio sensing [25]. [14] experimented with three variants of deep learning approaches such as feed-forward, convolutional, and recurrent neural networks on HAR datasets, and present guidelines for training neural networks. [34] proposed the DeepConvLSTM model in which convolutional layers extract the features from raw IMU data, and Long-Short Term Memory (LSTM) recurrent layers capture temporal dynamics of feature activations to improve the performance of HAR.

Deep neural networks have also helped applications that need to recognize hand gestures using surface electromyographic (sEMG) signals generated during muscle contractions [3, 24, 46]. [56] proposed a self-re-calibrating framework which can be updated to maintain the model's performance so that it does not need users' additional labels for re-training. [3] used sEMG of the forearm to classify finger touches with their proposed neural architecture combining convolutional, feed-forward, and LSTM layers.

Many works have investigated using deep learning for audio sensing tasks including Emotion Recognition, Speaker Identification [4], and Keyword Spotting. [11] proposed a deep learning modeling and optimization framework that specifically targets various audio sensing tasks in resource-constrained embedded systems. Keyword recognition [8] achieved 45% relative improvement with a deep learning model compared to a competitive Hidden Markov Model-based system.

In contrast to these works, we investigate whether current IL methods can enable a practical continual learning system for mobile and embedded sensing applications on-device and what the performance implications of such systems are. In addition, to fully understand the issue of CF in mobile sensing where the modality of the data is significantly different from image datasets [38] with which the IL methods are typically evaluated, we implement an end-to-end continual learning framework that evaluates various IL methods in three embedded sensing applications (e.g., HAR, GR, and ER) with different data modalities (e.g., accelerometer, sEMG, and speech).

## 3 CONTINUAL LEARNING FOR MOBILE AND EMBEDDED SENSING FRAMEWORK

We now present our framework to comprehensively evaluate the performance of various IL methods for three mobile and embedded applications (HAR, GR, and ER). We first explain the continual learning setup and three scenarios adopted in our experiments (§3.1). Then, we present six IL methods evaluated in this work (§3.2). We then describe the hyper-parameters of the LSTM based deep learning model and the different IL methods (§3.3). After that, we propose our novel IL model training process in §3.4. Next, we describe the datasets used in this study (§4.1). Finally, we provide brief details about our implementation (§3.5)

### 3.1 Continual Learning Setup and Three Scenarios

In this work, we focus on Sequential Learning Tasks (SLTs) from the mobile and embedded systems domain where new classes can emerge over time. Thus, the learning model has to continuously learn to accommodate new classes without CF, as would happen in real-life scenarios. Learning tasks of this type, called SLTs, indicates that a model continuously learns two or more tasks $D_1, ..., D_k$, one after another instead of learning a single task $D$ once [36]. Figure 1 shows an overview of our continual learning system for sensing applications using HAR as an illustrative example. A user starts with a model containing a fixed set of classes on their devices which is then incrementally updated over time as new classes arrives.

We introduce three scenarios of different levels of difficulties for models to learn continuously (from easy to difficult scenarios). First of all, inspired by Pfulb et al. [36], we adopt the SLTs consisting of two tasks: $D_1$ and $D_2$. Hence, Scenario 1 consists of two tasks, where the first task contains the $N - 1$ classes, and the second task contains the other one class ($N$ is the total number of classes). Scenario 2 includes two tasks where the first task contains half of the classes, $N/2$, and the second task contains the remainder of the classes. Finally, Scenario 3 deals with a more realistic situation where many tasks are to be learned sequentially [20]. In the third scenario, we first train a model in the first task with $N/2$ classes and then incrementally train the model by adding subsequent tasks with one class (essentially $N/2+1$ tasks). Unlike the first scenario (which has only N different cases of task permutations), it is not practical to consider every random permutation of classes to be included in different tasks for the second and third scenarios. Hence, we consider ten variations by randomly choosing classes in each task for the last two scenarios. Note that each task consists of disjoint groups of classes as we adopt class-incremental learning [50].

### 3.2 Incremental Learning Methods

As described in the related work section, various methods exist that can mitigate CF in IL. We describe them in depth as they form the basis of our exploration. To mitigate CF, there exist three

321

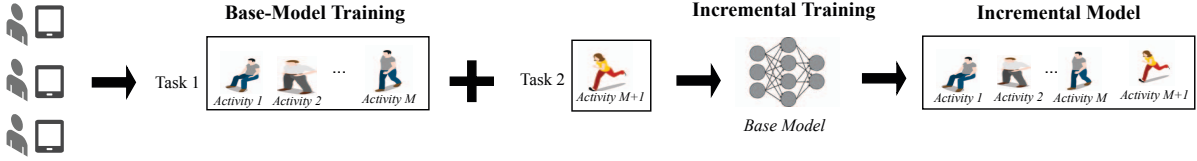

Figure 1: Overview of our continual learning system.

main categories of IL approaches: (1) Regularization, (2) Replay, and (3) Replay with Exemplars. We select at least one representative method for each of the above categories. These methods are the state of the art methods (most cited) for IL and are most often used in machine learning papers for comparison. We now describe the employed methods.

**LSTMs** [17]: LSTMs are a type of recurrent neural networks widely used for a sequence classifier in many applications, specifically for time-series data. We use LSTMs as a base neural network.

**EWC** [22]: Elastic Weight Consolidation (EWC) is a regularization based method which adds a penalty to regular loss function when learning a new task (i-th task), i.e.,

$$L(\theta) = L_i(\theta) + \lambda/2 \sum_{j=0}^{i-1} F_j (\Theta_i - \Theta_j^*)^2 \quad (1)$$

where $L(\theta)$ is the total loss, $\theta$ is the network's parameters, $L_i(\theta)$ is the loss for the new task, and $\Theta_j^*$ are the important parameters of all previous tasks. $\lambda$ is a hyperparameter that controls how much importance should be given to previous tasks compared to the new task. $F$ is the Fisher matrix used to constrain the parameters important to previously learned tasks to stay close to their old values to retain the knowledge of previous tasks and to be able to learn new tasks simultaneously.

**Online EWC** [43]: It is a variation of EWC method where the loss function is represented as,

$$L(\theta) = L_i(\theta) + \lambda/2 (\Theta_i - \Theta_{i-1}^*)^2 \sum_{j=0}^{i-1} F_j \quad (2)$$

Online EWC eliminates the need to store mean and fisher matrices for each previous task and only requires the latest mean and running sum of fisher matrices to calculate the current task's total loss.

**SI** [55]: It is another regularization method which is similar to EWC where the loss function is calculated in the following way,

$$L(\theta) = L_i(\theta) + \lambda \sum_k \Omega_k^i (\theta_k^* - \theta_k)^2 \quad (3)$$

where k is the subscript for the parameters of the models, $\lambda$ is the strength parameter, $\theta_k^*$ is the parameter value at the end of the previous task, and $\Omega_k^i$ represents the per-parameter regularization strength taking into account all previous tasks, calculated as:

$$\sum_{j=0}^{i-1} \frac{w_k^j}{(\triangle \theta_k^j)^2 + \varepsilon} \quad (4)$$

parameter distance $\triangle \theta_k^j$ determines how much a parameter moved between tasks during the entire trajectory of training. $\varepsilon$ is the dampening parameter to prevent division by zero errors. The main difference between SI and EWC is that SI weights importance, $w_k$, is continuously updated online during training. In contrast, in EWC, the Fisher matrices (weights importance) are calculated at the end of each task.

**LwF** [28]: This method relies on adding loss for the replayed data to the loss of the current task. The replayed data is the input data of the current task which is labeled using the model trained on the previous tasks to generate target probabilities. The ultimate aim of the replayed data is to match the probabilities predicted by the model being trained to the target probabilities (a form of data distillation) and is termed as the loss for replayed data.

**iCaRL** [40]: Incremental Classifier and Representation Learning (iCaRL) store data from previous tasks (i.e., exemplars) to alleviate the CF problem. The exemplars are a representative set of the small number of samples from a distribution, and those that can approximate the average feature vector over all training examples are selected as exemplars (based on herding [52]). The classification is done based on a nearest-class-mean (NCM) rule using features extracted from the deep learning model, where the class means are calculated from the stored examples. When new tasks (classes) arrive, iCaRL creates a new training set combining the exemplars from all the previous tasks with the data samples of the new task. Then, the model parameters are updated by minimizing a loss function which encourages the model to output the correct class for the new task (classification loss) and to reproduce the scores stored in the previous step for the old tasks (distillation loss) using data samples from the new training set.

**GEM** [29]: Gradient Episodic Memory (GEM) stores exemplars from the previous tasks like iCaRL and solves CF as a constrained optimization problem. A parameter update while doing IL is made depending on whether it will lead to an increase in loss for the previous tasks. This is calculated by computing the angle between loss gradient vectors of stored examples and the proposed parameter update. If the calculation suggests no loss, then the update is done straight away. Otherwise, the parameter is updated by projecting gradient in such a way that it will incur a minimal loss for the previous tasks.

**Our Contribution**: It is worth noting that the above six IL methods are known in the machine learning literature from a theoretical point of view. Yet, they are not off-the-shelf methods that can be simply used to any dataset to enable continual learning. As will be shown in Section 5, there exist many factors affecting the performance and applicability of the IL methods in real-world deployment such as the complexity of the continual learning scenario, resource availability of mobile and embedded devices, and choice of

322

hyper-parameters. Thus, a distinctive contribution of our work is a comprehensive evaluation and comparison study of the IL methods in diverse sensing applications and is to develop an end-to-end and on-device IL framework that can investigate trade-offs between performance, storage requirements, and latency.

### 3.3 Characterization of Hyper-parameters

We categorize hyper-parameters into three types and IL-method-specific parameters. First of all, we use architectural hyper-parameters which cannot be changed when learning new tasks, e.g., the number of hidden layers $L$ and its size $S$. We then use learning and regularization hyper-parameters which can be adaptable when learning new tasks. For example, a learning rate $\epsilon$ and $\lambda$ term in L2-regularization can be modified during training over time. We denote the set of hyper-parameters as $\mathcal{P}$.

**IL-method-specific parameters:** Each IL method has method-specific parameters to control the behaviors of the model. For example, in regularization-based methods [22, 28, 44, 55], importance parameter $\lambda$ is often utilized to modulate how much importance a model puts on previous tasks or a current task. The importance parameter can be adaptable while learning new tasks in our IL model training process (Algorithm 1). In addition, in replay with exemplars-based methods [29, 40], the size of the storage budget is used to balance between storage requirements and the performance of a model. Since the budget size is difficult to be adaptable after completion of the first task, it is given as an input in our experimental protocol (Algorithm 1).

**Hyper-parameter setting for experiments:** We first fix several hyper-parameters as default values. We set dropout rates for all tasks as 0.2 and 0.5 in input and hidden layers of a model, respectively [16] and a batch size of 32 with Adam optimizer set to a default learning rate of 0.001 for task 1 ($D_1$). After that, we vary hyper-parameters for all models in each dataset. Specifically, in the task 1 ($D_1$), we vary architectural hyper-parameters as follows: $L \subset \{1, 2\}$, $S \subset \{32, 64\}$. In subsequent tasks from task 2 to k ($D_2, ..., D_k$), we fix architectural hyper-parameters but vary adaptable hyper-parameters and IL-method-specific parameters as follows: (1) $\epsilon \subset \{0.001, 0.0001\}$ for all models, (2) $\lambda \subset \{1, 10, 10^2, 10^3, 10^4, 10^5, 10^6\}$ for both EWC and Online EWC, (3) $\gamma \subset \{0.5, 1.0\}$ for Online EWC, (4) $c \subset \{0.2, 0.4, 0.6, 0.8, 1.0\}$ for SI. We denote varying IL-method-specific parameters as $\mathcal{P}_{IL}$. For replay-based methods, the losses of the current and replayed data are weighted according to the number of tasks a model has learned so far by following [50]. Note that budget size, $\mathcal{B} \subset \{1\%, 5\%, 10\%, 20\%\}$, is given as an input and fixed for replay with exemplars-based methods while other hyper-parameters are permuted. Since the total number of samples for each dataset is different, we use a ratio from the total training samples rather than a fixed number of samples for the budget size.

### 3.4 Model Training Process

We extend protocol [36] to incorporate multiple tasks up to task k ($D_1, ..., D_k$) in an incremental manner based on our characterization of hyper-parameters and IL-method-specific parameters. Algorithm 1 describes our protocol in which we only utilize training data of a current task j ($\le k$) for model learning and test data of previously learned tasks up to task j for evaluation.

---

**Algorithm 1:** IL model training process to determine the best model by incrementally learning tasks up to task k

**Input:** Tasks $D_1, ..., D_k$, model $m$, budget $\mathcal{B}$, epochs $\mathcal{E}$
**Input:** The number of hidden layers $L$, Hidden layer size $S$
**Input:** IL-method-specific parameters $\mathcal{P}_{IL}$, learning rate $\epsilon$
**Output:** The best model with hyper-parameter vector $p^*$

1 **for** $p \in (L \cup S)$ **do**
2     **for** $t = 1, \mathcal{E}$ **do**
3         Train model $m_1$ using training set of $D_1$ with $p$
4         Test model $m_1$ using test set of $D_1$
5         Store performance $q_{1,t}$
6 Update the model $m_{1,p^*}$ with max $q_1$
7 **for** $p \in (\mathcal{P}_{IL} \cup \epsilon)$ **do**
8     Initialize model $m_2$ with $m_{1,p^*}$
9     **for** $j = 2, k$ **do**
10         **for** $t = 1, \mathcal{E}$ **do**
11             Train model $m_2$ using training set of $D_j$ with $p$
12             Test model $m_2$ using test set of $\cup_{l=1}^{j} D_l$
13             Store performance $q_{j,t}$
14 Update the model $m_{k,p^*}$ with max $q_k$

---

Given an SLT consisting of $D_1, D_2, ..., D_k$ and a model $m$, the goal is to find a vector of hyper-parameters $p^*$ which produces the best performance $q$ after incrementally training all tasks up to task k. For the first step, we find the best performing hyper-parameters in task 1 ($D_1$) by searching among the set of architectural hyper-parameters (lines 1-5 in Algorithm 1) and update the model $m_{p^*}$ with the found hyper-parameters (line 6). The next step is to find the best model by searching among the set of learning hyper-parameters and IL-method-specific parameters in subsequent tasks from task 2 to k (lines 7-13). Finally, we select the best model which shows the highest performance based on test sets after incrementally trained up to task k (line 14). Note that to facilitate the extensive experiments performed in our study and to make a fair comparison among the IL methods (Section 5), we first identify the best architectural hyper-parameter (from $L \subset \{1, 2\}$ and $S \subset \{32, 64\}$) and then use the found hyper-parameter across the different IL methods. The final LSTM architecture we used for each dataset are reported in Table 1.

### 3.5 Implementation

We implemented our continual learning framework on Nvidia Jetson Nano and One Plus Pro smartphone platforms. All the IL algorithms were explored on Nano GPU, and we used PyTorch 1.1 to implement the framework. Keeping in mind that Scenario 3 is the most practical continual learning scenario and iCaRL is the best performing IL approach, we only implemented iCaRL for Scenario 3 on the smartphone's CPU (as an Android app) using the DeepLearning4j library. The smartphone app size is 134 MB. We choose CPU on the smartphone as it provides an upper bound on the performance of any system and is more challenging to implement. We envisage that if a system can work (or at least feasible) on a CPU, then it would be much easier and faster to run similar systems on accelerators such as GPU. When working on a dataset, we first loaded the training data pertaining to all the tasks in the memory to make the continual learning process work faster. As a limited amount of memory is allocated to each Android app, we set large heap property in the app to True to use larger heaps for our app. We still encountered memory issues, especially when working with large datasets such as Skoda, which we solved by using memory-mapped files.

In addition, we employ a weighted F1-score which is more resilient to class imbalances as the employed datasets (see §4.1 for details) are not balanced [14, 49]. As in [21], we applied a weighted loss to all evaluated methods by estimating the inverse class distribution which gives more importance to the loss of a class with fewer samples. Also, as deep learning models can overfit to small datasets such as EmotionSense, with our framework, we experimented with shallow and deep neural network architectures and found that deep architectures show marginal improvement over shallow architectures, indicating that the overfitting is not an issue.

## 4 EXPERIMENTAL SETUP

Before we present the findings of this work in Section 5, we describe experimental setup for conducting a comprehensive evaluation of three continual learning schemes in mobile and embedded sensing applications. We first describe six datasets in three different sensing applications (§4.1) and evaluation metrics adopted for systematic comparison of the IL techniques and their trade-offs between system aspects (e.g., storage and computational costs) (§4.2).

### 4.1 Datasets

We focus on three sensing applications (e.g., HAR, GR, and ER) as they are some of the most popular applications in the mobile sensing. Table 1 shows the overview of the employed datasets.

*4.1.1 Human Activity Recognition (HAR).* For the HAR application, we used three datasets: (1) HHAR [49], (2) PAMAP2 [41], and (3) Skoda [48]. These datasets contain many real-life activities (e.g., walking, sitting, and cycling) obtained using Inertial Motion Units (IMUs), which contain accelerometer, gyroscope, and magnetometer data of mobile and wearable devices. We next present the detailed summaries of the three datasets.

**HHAR:** This dataset considers six different daily activities of users The data was recorded from nine participants, where they followed a scripted set of activities with eight smartphones and four smartwatches of different brands and models. Having various devices for recording makes HHAR an excellent benchmark to study heterogeneity of HAR (i.e., sensor biases, sampling rate heterogeneity, and sampling rate instability). We follow the preprocessing steps as proposed by Yao et al. [53]. Raw measurements of both accelerometer and gyroscope are segmented into 5-second samples. Each sample is divided into time intervals of 0.25s. After that, we apply a Fourier transform to each time interval. It produces $d \times 2f$ dimensional vectors per time interval, where $d$ is the dimension for each measurement and $f$ is the frequency with magnitude and phase pairs, resulting in 120 dimensions. We adopt leave-one-user-out (LOUO) for evaluation [53]. One user (i.e., the first participant) is used for testing, and the remaining users are left for training.

**PAMAP2:** In this dataset, nine subjects carried out various daily living activities and sportive exercises. IMU data (accelerometer, gyroscope, magnetometer), heart rate, and temperature data were recorded from body-worn sensors attached to the hand, chest, and ankle. The resulting dataset has 52 dimensions, and more than 10 hours of data were collected. We follow a preprocessing protocol used by Hammerla et al. [14]. The sensor data are downsampled to 33Hz. After that, all samples are normalized to zero mean and unit variance. Also, to be consistent with the previous works [13–15], we use runs 1 and 2 from the sixth participant for testing and remaining data for training.

**Skoda:** The Skoda dataset contains activities of assembly-line workers in a manufacturing scenario. One subject wore 20 3D accelerometers on both arms. Following the preprocessing steps [13, 34], we employ raw and calibrated data from ten accelerometers placed on the right arm, resulting in input data of 60 dimensions. The data are downsampled to 33Hz and normalized to zero mean and unit variance. For experiments, the last 10% of each class is used as the test data and the remaining as the training data. Note that Skoda consists of one subject, i.e., subject dependent evaluation.

*4.1.2 Gesture Recognition (GR).* We employ the Non Invasive Adaptive Prosthetics (Ninapro) database [2] for the GR application in our experiments as it consists of surface electromyography (sEMG) signals and thus can provide different sensor modalities than IMU sensors present in HAR datasets.

**Ninapro (Per Subject):** The Ninapro database is widely used in research on the hand movement recognition application. We employ Ninapro Database2 (DB2) in this study. It includes sEMG data recordings from 40 subjects while performing several repetitive gestures such as wrist movements, grasping and functional movements, and force patterns. Following, Li et al. [27], we select ten types of hand gestures commonly used in daily life. After that, we downsample the sEMG data to 200 Hz and normalize them to zero mean and unit variance. We used a sliding window size of 200 ms with a 50% overlap [7, 56]. We select a subject who has the most amount of data samples for subject dependent (i.e., per subject) evaluation. After that, we use the fifth repetition for a test set and the remainder for training.

**Ninapro (LOUO):** To have consistent evaluation with the HAR application we adopt LOUO evaluation for the GR application using the Ninapro dataset. We select the top ten subjects having more data samples than others. After that, we use a subject with the least data samples for testing and the remainders for training. The preprocessing steps are the same as in Ninapro (Per Subject).

*4.1.3 Audio Sensing Task.* We pick Emotion Recognition (ER) since it is one of the most widely adopted audio sensing tasks. We employ the EmotionSense dataset [39] which was collected by recording human participants' emotions as well as proximity and patterns of conversation using an off-the-shelf smartphone. This dataset has been used in multiple studies to understand the correlation and impact of interactions and activities on the emotions and behavior of individuals in various settings [12][25][30].

**EmotionSense:** The EmotionSense dataset contains audio signals which represent 14 different emotions. In the EmotionSense dataset, each measurement corresponding to a particular emotion (or class) is based on a 5-second context window. Following Georgiev et al. [11], we extract 24 log filter banks [47] from each audio frame over a time window of 30 ms with 10 ms stride. Each

324

Table 1: Overview of the employed datasets.

| Application | Dataset | Dimension | # Train Data | # Test Data | # Classes | Layer/Size |
|---|---|---|---|---|---|---|
| HAR | HHAR | 20 × 120 | 59,403 | 7,721 | 6 | 2/64 |
| | PAMAP2 | 33 × 52 | 35,263 | 5,209 | 12 | 1/64 |
| | Skoda | 33 × 60 | 10,047 | 1,193 | 10 | 1/64 |
| GR | Ninapro (Per Subject) | 40 × 12 | 3,118 | 639 | 10 | 1/64 |
| | Ninapro (LOUO) | 40 × 12 | 30,488 | 3,759 | 10 | 1/64 |
| ER | EmotionSense | 20 × 24 | 2,011 | 224 | 14 | 2/64 |

sample contains 500*24 = 12,000 features where 1–24 features are filter banks from the first 10 ms, and 25–48 features are filter banks for the next 10 ms and so on. After that, as our preprocessing steps, we downsample each sample measurement by averaging corresponding 24 filter banks of every 250 ms (or 25 consecutive windows) without any overlap to reduce the length of the input sequence for a learned neural network. We normalize each window to zero mean and unit variance.

### 4.2 Evaluation Metrics

We consider how much an IL method forgets previous tasks and learns new tasks after it was trained from task 1 to k to assess the actual performance of IL methods [6] by considering the following metrics.

**Average Performance Measure (A)**: We denote the performance measure of a model on the j-th task ($j \leq k$) as $a_{k,j} \in [0,1]$ after the model is trained from task 1 to k. The average performance measure at task k is defined as follows:

$$A_k = \frac{1}{k} \sum_{j=1}^{k} a_{k,j} \quad (5)$$

The output space consists of $\cup_{j=1}^{k} y^j$, and $a_{k,j}$ is based on a weighted F1-score in this work. Note that $a_{k,j}$ can be used to indicate an accuracy, proportion of correctly classified activities or gestures.

**Forgetting Measure (F)**: The forgetting measure provides an estimate of how much a model forgets about the task given its present state. The forgetting for the j-th task after the model has been trained up to task $k > j$ can be quantified as:

$$f_j^k = \max_{l \in 1,\ldots,k-1} a_{l,j} - a_{k,j}, \quad \forall j < k \quad (6)$$

The average forgetting at k-th task is denoted as $F_k = \frac{1}{k-1} \sum_{j=1}^{k-1} f_j^k$ by normalizing the number of tasks seen previously. The lower the $F_k$, the less forgetting on previous tasks.

**Intransigence Measure (I)**: Intransigence is defined as the inability of a model to learn new tasks. To quantify the inability to learn, the joint model, often considered upper bound, which has access to all the datasets seen so far ($\cup_{l=1}^{k} D_l$) is compared and its performance is denoted as $a_k^*$. We then denote the intransigence for the k-th task as:

$$I_k = a_k^* - a_{k,k} \quad (7)$$

where $a_{k,k}$ represents the performance of a model on the k-th task trained up to task k. Lower $I_k$ implies that a model performs as close as a joint model or performs even better than the joint model when intransigence is negative ($I_k < 0$). Note that we use $a_{k,k}$ and $I_k$ as the main performance indicators of a model since we are interested in the current performance of the model on all learned tasks from 1 to k.

Note that in addition to metrics mentioned above, we also report **storage** and **latency** required to execute each IL method.

## 5 FINDINGS

We now present the results of our evaluation. Firstly, we compare the performances of different IL methods on HAR, GR, and ER tasks using two basic scenarios (Scenario 1 and 2) in §5.1. Then, we study the performance of IL methods for Scenario 3 in §5.2. We examine the generalizability of IL methods across different datasets (§5.3). Then, we discuss the trade-offs of IL methods with respect to the storage, computational costs, and memory footprint. (§5.4). Finally, in §5.5, we investigate the effect of iCaRL specific parameters on the performance.

### 5.1 Performance on Simple and Mildly Difficult Tasks

We show the best average weighted F1-scores across all runs for different IL methods for Scenario 1 and 2 for different datasets in Figure 2 and Figure 3, respectively. For HAR and GR applications, the results of iCaRL/GEM with the budget size of 20% are shown in Figure 2 and Figure 3 since the models with the budget size of 20% show the best performance. Then, for the ER application, the results of iCaRL/GEM with the budget size of 40% are shown since the EmotionSense dataset has the least number of training samples, requiring more budget size in ER than the other two applications. **Joint** refers to the case when training data is available for all the classes from the beginning. It is a classic case to train a model with all data at once and serves as the upper bound in many cases. **None** refers to the case when no IL method is applied to solve CF. The white part in the figure shows performance on Task 1, and the grey part shows performance for Task 2.

The results show that without any IL method (None), the performance drops sharply as soon as a new task is encountered. The decline in performance is as drastic as 60% in both scenarios. iCaRL provides the best performance in Scenario 1, which stays very close to the performance obtained with the joint model. It is because iCaRL stores representative exemplars and relies on a nearest-class-mean (NCM) rule that is robust against changes in the data representation [40]. In fact, all the IL methods effectively solve the CF problem and achieve comparable performance to the joint model (between 5% and 15%) after only running for a few epochs (5 or less in many cases). *One can conclude that, in general, the existing IL*

325

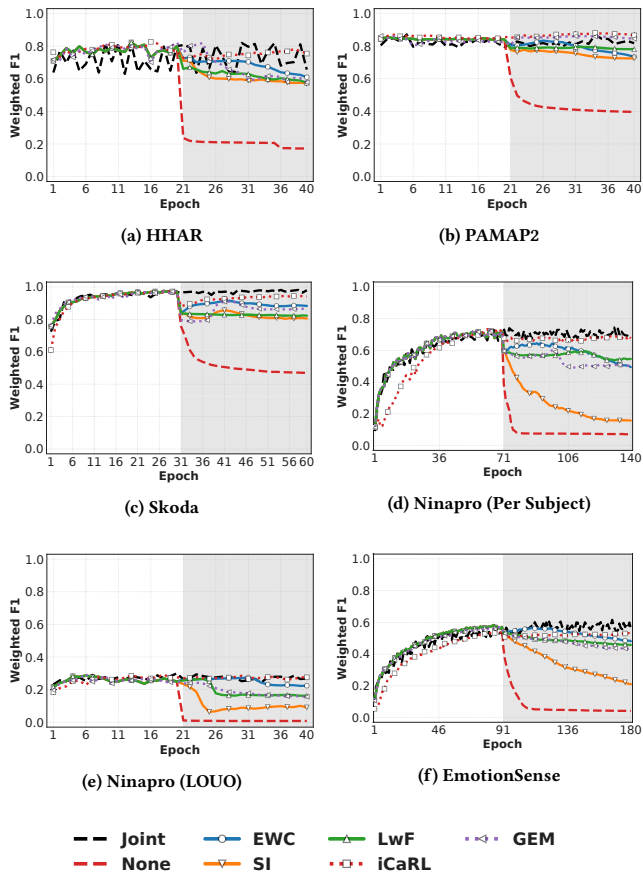

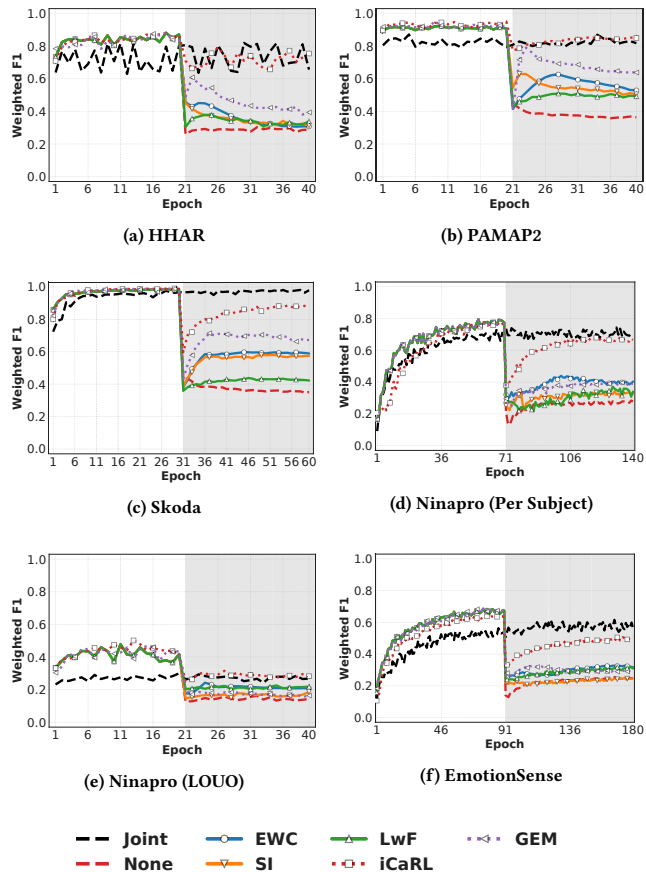

Figure 2: The performance comparison of the five IL methods including two baselines in Scenario 1 on each dataset.

Figure 3: Performance comparison in Scenario 2.

methods we analyzed can solve the CF issue on mobile and embedded sensing applications for simple scenarios.

However, the same cannot be said for the performance in Scenario 2. *Except iCaRL, none of the other methods seems to solve the CF issue for the mildly complex scenario (i.e., Scenario 2).* The performance drop is up to 60% when the performances between IL methods and the joint model are compared. iCaRL remains the best performing method with its weighted F1-score close to that of the joint model (within 10%). GEM performs the second-best (within few epochs) on HAR datasets while EWC performs well for GR and ER datasets. Although GEM is a replay with exemplars-based approach like iCaRL, it never matches the performance of iCaRL due to its reliance on using gradients and not the actual examples themselves. Another reason might be that iCaRL selects best examples to be stored based on herding (a sort of prioritization), while GEM employs selecting examples randomly which can be less informative. A regularization-based method such as SI and a replay only approach such as LwF perform poorly across all datasets. The weighted F1-score degrades roughly 40–50% of what can be achieved by the joint model. As indicated by [1], the performance of LwF significantly decreases when the model learns a sequence of tasks drawn from different distributions. In other words, when tasks learned by LwF are not sufficiently related, enforcing the new model to give similar outputs for the old task may hurt the model's performance. SI relies on the weight changes in a batch gradient descent which can overestimate the importance of the weights and thereby leads to lower performance.

Note that iCaRL employs a different way (i.e., NCM rule) to classify data samples (perform inference) than other methods (including None and Joint) which use cross-entropy based classification. Also, for GEM, it minimizes the loss on the current task by using inequality constraints, avoiding its increase but allowing its decrease. Therefore, iCaRL and GEM can obtain different weighted F1-scores than the other methods in task 1. Otherwise, ideally one would assume all methods (e.g., None, EWC, SI, LwF in our study) to get the same performance in the first task as it only involves learning a baseline LSTM model without any IL. Also worth mentioning is that initially (especially task 1) IL methods can achieve higher weighted F1-scores than the joint model. It is because their performance is based on classifying the smaller number of classes than the joint model, where all classes need to be classified from the first epoch.

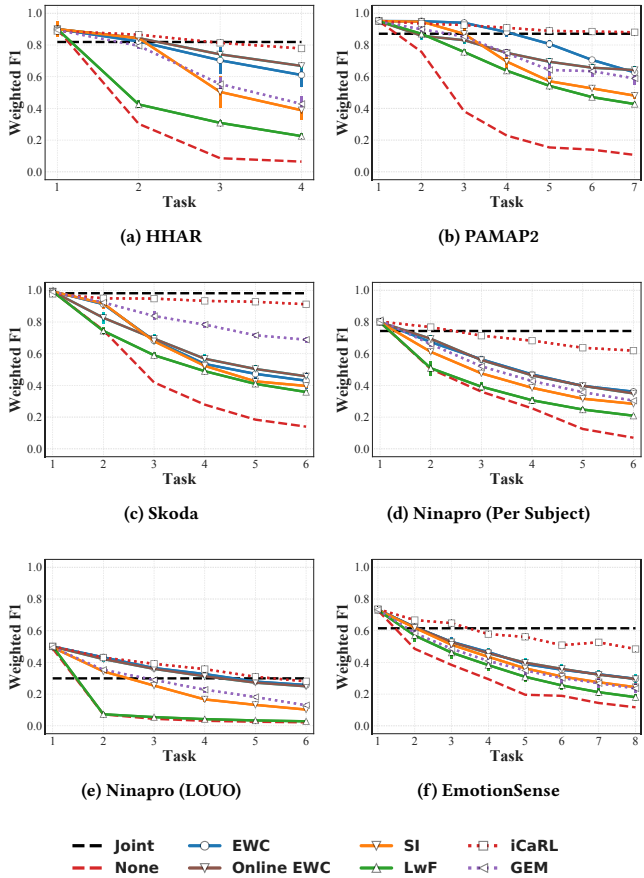

(a) HHAR (b) PAMAP2 (c) Skoda (d) Ninapro (Per Subject) (e) Ninapro (LOUO) (f) EmotionSense

Legend: Joint, EWC, SI, iCaRL, None, Online EWC, LwF, GEM

**Figure 4: The performance comparison in Scenario 3. All reported results are averaged over 10 trials, and standard-error intervals are depicted.**

## 5.2 Performance on Many Sequential Tasks

Figure 4 shows results for Scenario 3. Recall that Scenario 3 presents the case when classes are added one by one to an already existing deep learning model, which will happen in real-life scenarios and is the most challenging task for any IL method. Note that this graph is shown differently than the graphs for Scenario 1 and 2 (epoch based) as in epoch based graph, we would have only two data points to show as there were only two tasks. In Scenario 3 the number of tasks will be $N/2 + 1$ for N classes. Without the IL method (None), CF happens, and the weighted F1-score almost always lies between 0%–10%. *iCaRL is the best method and appears to solve the CF issue for the challenging third scenario*. Its performance is nearly equal to the joint model in most of the cases. *All other methods do not solve the CF issue, and the performance suffers severely as more tasks are added to the system especially with LwF and SI.*

**Table 2: Average performance of different methods in all scenarios on HAR, GR, and ER.**

| Scenario | Methods | HAR | | | | GR | | | | ER | | | |
|---|---|---|---|---|---|---|---|---|---|---|---|---|---|
| | | $A_k$ | $F_k$ | $a_{k,k}$ | $I_k$ | $A_k$ | $F_k$ | $a_{k,k}$ | $I_k$ | $A_k$ | $F_k$ | $a_{k,k}$ | $I_k$ |
| 1 | None | 0.55 | 0.35 | 0.54 | 0.36 | 0.22 | 0.29 | 0.20 | 0.32 | 0.47 | 0.11 | 0.44 | 0.16 |
| | EWC | 0.88 | 0.01 | 0.86 | 0.03 | 0.49 | 0.01 | 0.47 | 0.05 | 0.60 | 0.01 | 0.58 | 0.03 |
| | SI | 0.85 | 0.03 | 0.81 | 0.07 | 0.45 | 0.04 | 0.42 | 0.10 | 0.57 | 0.05 | 0.54 | 0.07 |
| | LwF | 0.84 | 0.02 | 0.79 | 0.10 | 0.47 | 0.02 | 0.44 | 0.09 | 0.57 | 0.01 | 0.54 | 0.07 |
| | iCaRL | 0.89 | 0.01 | 0.88 | 0.01 | 0.51 | 0.01 | 0.49 | 0.03 | 0.57 | 0.02 | 0.56 | 0.05 |
| | GEM | 0.88 | 0.01 | 0.87 | 0.02 | 0.46 | 0.03 | 0.43 | 0.09 | 0.57 | 0.01 | 0.54 | 0.06 |
| 2 | None | 0.30 | 0.76 | 0.41 | 0.48 | 0.20 | 0.48 | 0.23 | 0.29 | 0.27 | 0.45 | 0.27 | 0.34 |
| | EWC | 0.77 | 0.06 | 0.65 | 0.24 | 0.47 | 0.06 | 0.35 | 0.17 | 0.55 | 0.01 | 0.39 | 0.22 |
| | SI | 0.64 | 0.26 | 0.60 | 0.29 | 0.31 | 0.31 | 0.29 | 0.23 | 0.38 | 0.27 | 0.32 | 0.29 |
| | LwF | 0.70 | 0.03 | 0.48 | 0.41 | 0.45 | 0.06 | 0.31 | 0.21 | 0.52 | 0.04 | 0.35 | 0.26 |
| | iCaRL | 0.89 | 0.05 | 0.86 | 0.03 | 0.53 | 0.09 | 0.51 | 0.02 | 0.57 | 0.07 | 0.53 | 0.08 |
| | GEM | 0.77 | 0.13 | 0.71 | 0.18 | 0.39 | 0.19 | 0.31 | 0.21 | 0.51 | 0.08 | 0.37 | 0.24 |
| 3 | None | 0.22 | 0.21 | 0.10 | 0.79 | 0.09 | 0.17 | 0.05 | 0.48 | 0.18 | 0.16 | 0.12 | 0.49 |
| | EWC | 0.75 | 0.01 | 0.56 | 0.34 | 0.44 | 0.01 | 0.31 | 0.21 | 0.46 | 0.01 | 0.30 | 0.49 |
| | Online EWC | 0.72 | 0.03 | 0.59 | 0.30 | 0.44 | 0.01 | 0.30 | 0.22 | 0.45 | 0.01 | 0.30 | 0.31 |
| | SI | 0.59 | 0.10 | 0.42 | 0.47 | 0.32 | 0.07 | 0.22 | 0.31 | 0.42 | 0.02 | 0.24 | 0.36 |
| | LwF | 0.53 | 0.06 | 0.34 | 0.55 | 0.20 | 0.08 | 0.12 | 0.40 | 0.29 | 0.11 | 0.18 | 0.43 |
| | iCaRL | 0.86 | 0.01 | 0.79 | 0.10 | 0.53 | 0.01 | 0.45 | 0.07 | 0.62 | 0.12 | 0.48 | 0.13 |
| | GEM | 0.70 | 0.07 | 0.57 | 0.32 | 0.33 | 0.08 | 0.22 | 0.31 | 0.33 | 0.02 | 0.16 | 0.44 |
| - | Joint | - | - | 0.89 | - | - | - | 0.52 | - | - | - | 0.61 | - |

### 5.3 Generalization

Table 2 shows the results in a summarized way for all the datasets and IL methods evaluated in our study. $A_k$ refers to average performance on all tasks while $a_{k,k}$ shows the weighted F1-score at the end of learning all tasks. $F_k$ tells us how good an IL method is in retaining old knowledge about previous tasks. Whereas $I_k$ means how much an IL method is good at learning new tasks. Note that the higher the values of $A_k$ and $a_{k,k}$, the better the model is. However, for $F_k$ and $I_k$, a low value indicates a better model since low $F_k$ and $I_k$ means that the model forgets knowledge of previous tasks less and performs as close as a joint model, respectively. iCaRL is one of the best-performing methods on all metrics across all datasets. iCaRL can learn new classes (tasks) while retaining old knowledge and maintain high performance even in the most challenging scenario. Given that small errors are allowed when performing HAR, GR and ER, iCaRL alleviates the issue of CF to a large extent. The same is not true for all other IL methods. Although LwF allows previous knowledge to be largely retained (low F value), it does not learn new tasks easily and thus has low performance in general. SI is neither good at learning new tasks (high I) nor at remembering old knowledge (high F). EWC and online EWC offer a decent alternative to iCaRL without needing extra storage on-device but at the expense of lower performance than iCaRL. The overall takeaway is that *iCaRL can enable a system to learn incrementally (continuously) in the mobile and embedded sensing domain (if storage is not such a constraint on a device).*

### 5.4 Storage, Latency, and Memory Footprint

**Storage:** We report the storage overhead of each IL method, as shown in Table 3. We first specify the mathematical formulas used to calculate the overall storage requirements of each IL method to show how much storage the IL method needs with respect to the number of tasks ($\mathcal{T}$) added, the model parameters ($M$), and the budget size ($\mathcal{B}$). This point would help practitioners and researchers easily understand how much storage overhead occurs when they want to deploy their models with a particular IL method. First of all, LwF requires no extra storage other than the storage needed to store the model parameters ($M$). Then, SI requires a running estimate ($w_k$), the cumulative importance measures ($\Omega_k^i$), and reference weights ($\theta_k^*$) of importance weights of the current task. EWC stores fisher matrices and means for each task. Unlike EWC, Online EWC is only required to store one fisher matrix and running means across tasks. Thus, the required storage for Online EWC does not increase as the number of learned tasks increases. Similar to LwF, iCaRL also requires the previous task model for knowledge distillation. For GEM, it stores the gradient of the exemplar set for each learned task. As both iCaRL and GEM rely on stored examples, their storage demands are mainly driven by the number of examples to be stored (i.e., budget size, $\mathcal{B}$).

Table 3: Storage requirements of IL methods. $\mathcal{M}$ refers to the number of model parameters, $\mathcal{T}$ represents number of tasks and $\mathcal{B}$ is the storage budget.

| Category | Method | Required Storage |
|---|---|---|
| Reg-based | EWC | $2 \times \mathcal{M} \times \mathcal{T}$ |
|  | Online EWC | $2 \times \mathcal{M}$ |
|  | SI | $3 \times \mathcal{M}$ |
| Replay-based | LwF | $\mathcal{M}$ |
| Replay+Exemplars | iCaRL | $\mathcal{M} + \mathcal{B}$ |
|  | GEM | $\mathcal{T} \times \mathcal{M} + \mathcal{B}$ |

Numerical model sizes (i.e., $\mathcal{M} + \mathcal{B}$) are shown in Table 4 for all the employed datasets in Scenario 3. Note that we do not add tables containing the results of Scenario1 and 2 due to the page limit. However, by reporting the results of Scenario 3 where the storage requirements of various IL methods are greater than or equal to those of Scenario 1 and 2, we aim to present the upper bound of the required storage. Besides, the reported numerical sizes of storage requirements in Table 4 are based on IL methods with the largest model in our experiments (i.e., number of LSTM layers ($L = 2$) and the number of hidden units ($S = 64$)) to capture the upper bound to practically operate IL methods on embedded and mobile devices. Here we take the Skoda dataset to further explain our findings as it represents an ideal use case scenario where IL methods need to be applied to personal mobile devices (single-user scenario with modest dataset size). In the Skoda dataset, replay with exemplars methods such as iCaRL and GEM requires at most around 17 MB, and other IL methods have even smaller storage requirements. For EmotionSense dataset where we use up to 40% budget, iCaRL needs less than 2 MB, and GEM needs less than 3.4 MB at most. Even with the largest dataset of HHAR in our experiments, the storage requirements are constrained within less than about 115 MB, which falls well within the storage capacity of modern embedded devices and smartphones. Many modern mobile and embedded devices already support a large amount of storage (in order of GBs).

*In summary, the amount of storage required to practically enable continual learning on many modern edge platforms such as Nvidia Jetson or Raspberry PIs and smartphones is not excessive, as evident from Table 4.* Note that tuning appropriate parameters in the IL method would still allow IL to perform effectively, i.e., ensuring good performance with a reasonable budget size (discussed in §5.5).

**Latency:** The average training and incremental learning time to execute different IL methods are illustrated in Table 5 for all the employed datasets in Scenario 3 on Jetson Nano[2] which is an edge platform having four cores, 4 GB RAM and a GPU and often used in mobile robotics and can be used in tablets. Training time represents the usual training time involved in learning a neural network including updating weights, back-propagation, etc. GEM is computationally the most expensive. On small datasets of Ninapro (Per Subject) and EmotionSense, IL time is around 57.3-85.2 seconds. Then, on the largest dataset of HHAR, IL takes up to 2,660 seconds. It is because gradient computation over previous tasks is computationally expensive. Also, EWC and Online EWC show high IL time, taking over 1,213 seconds in HHAR. This is surprising as EWC is a simple method. However, the time complexity comes from calculating and updating the Fisher matrices, which is a computationally expensive process, after every task. SI (mostly relying on running estimates) and LwF (replay only, calculating distillation loss) are two of the top three fastest IL methods but come at the peril of very low accuracy, making them unsuitable for IL in mobile and embedded applications. iCaRL, the best performing IL method, is also very fast and takes only a few seconds (e.g., 8.46–16.5 seconds) in the Ninapro (Per Subject) and EmotionSense datasets to complete. In the HHAR dataset, the average latency of IL time of iCaRL with the largest budget size (i.e., 20%) is relatively small of 150 seconds compared to its training time (i.e., 924 seconds) and the IL time of EWC (i.e., 1,213 seconds) and GEM (i.e., 2,660 seconds). In reality, most of the time is taken by actual training (except EWC and Online-EWC), which depends on the number of epochs to be performed and is independent of the IL method. Across scenarios, we observe that the average training time can range from one to 15 minutes in general (except GEM).

Having realized that iCaRL is the most promising method in terms of accuracy and latency, we wanted to check if iCaRL can also effectively work on modern smartphone CPUs. For this, we have implemented iCaRL on OnePlus 7 Pro for three datasets: Skoda, Ninapro (Per Subject), and EmotionSense as they represent datasets where IL needs to be applied to personal mobile devices (single-user case) and Scenario 3 (most practical scenario). The smartphone has eight cores and 12 GB of RAM. To reiterate, we used DeepLearning4j library to implement iCaRL. The smartphone app size is 134 MB. The results are shown in Table 6. Similar to Jetson Nano, iCaRL takes minimal time (0.5–212 seconds) for all the tasks for every dataset. This does not only mean that IL is feasible on modern smartphones but even if a very high number of tasks are to be learned even in the most challenging scenario, iCaRL can do end-to-end IL in a few minutes. The training time slows down the whole process and ranges from 20–75 minutes on the CPU of the smartphone for different datasets. Also note that the training time taken by the tasks after the first task (actual incremental tasks after the initial model is trained) is very small: one to four minutes. This is a relevant result as one can train a baseline model on a powerful machine first and can then move it to a mobile and embedded device to learn incrementally over time. Regardless, we show that the complete incremental learning process can still be done entirely on the smartphone CPU, especially given that the phone can be charged overnight. *This is an interesting result as this suggests that our continual learning framework can be deployed on a smartphone CPU. It is also encouraging because the performance can be further improved by exploiting GPU and NPU once support for training them programmatically starts to emerge.*

---

[2] By reporting the results of Scenario 3 where the latency of IL methods is greater than or equal to that of Scenarios 1 and 2, we aim to capture the upper bound of the latency.

Table 4: Storage requirements of IL methods for all datasets - Scenario 3. Units are measured in MB.

| IL Method | HHAR | PAMAP2 | Skoda | Ninapro (Per Subject) | Ninapro (LOUO) | EmotionSense |
|---|---|---|---|---|---|---|
| EWC | 2.601 | 3.599 | 3.177 | 2.587 | 2.587 | 3.663 |
| Online EWC | 0.650 | 0.514 | 0.529 | 0.431 | 0.431 | 0.458 |
| SI | 0.975 | 0.771 | 0.794 | 0.647 | 0.647 | 0.687 |
| LwF | 0.325 | 0.257 | 0.265 | 0.216 | 0.216 | 0.229 |
| iCaRL (1%) | 5.990 | 2.676 | 1.051 | 0.270 | 0.805 | 0.257 |
| iCaRL (5%) | 28.838 | 12.341 | 4.187 | 0.512 | 3.190 | 0.407 |
| iCaRL (10%) | 57.350 | 24.421 | 8.179 | 0.805 | 6.179 | 0.607 |
| iCaRL (20%) | 114.374 | 48.658 | 16.162 | 1.410 | 12.141 | 0.981 |
| iCaRL (40%) | - | - | - | - | - | 1.755 |
| GEM (1%) | 6.989 | 4.205 | 2.350 | 1.351 | 1.884 | 1.862 |
| GEM (5%) | 29.817 | 13.874 | 5.537 | 1.583 | 4.278 | 2.016 |
| GEM (10%) | 58.372 | 25.996 | 9.532 | 1.884 | 7.274 | 2.217 |
| GEM (20%) | 115.444 | 50.240 | 17.476 | 2.485 | 13.266 | 2.603 |
| GEM (40%) | - | - | - | - | - | 3.374 |

Table 5: Average Latency (Training Time/IL Time) in seconds for IL methods on different datasets - Scenario 3 on Jetson Nano.

| IL Method | HHAR | PAMAP2 | Skoda | Ninapro (Per Subject) | Ninapro (LOUO) | EmotionSense |
|---|---|---|---|---|---|---|
| EWC | 672/1213 | 329/599 | 120/170 | 173/73.1 | 251/558 | 159/67.8 |
| Online-EWC | 651/1188 | 291/570 | 105/162 | 148/60.2 | 225/539 | 131/46.3 |
| SI | 717/144 | 336/55.3 | 118/18.0 | 146/22.1 | 269/47.1 | 123/22.4 |
| LwF | 660/88.6 | 362/70.0 | 113/15.7 | 150/19.4 | 284/58.5 | 128/14.2 |
| iCaRL (1%) | 906/76.2 | 268/36.3 | 113/13.6 | 141/12.9 | 265/32.4 | 117/8.46 |
| iCaRL (5%) | 928/93.0 | 269/44.2 | 131/16.6 | 147/15.1 | 244/38.3 | 118/8.80 |
| iCaRL (10%) | 896/109 | 302/54.7 | 149/19.3 | 130/13.2 | 235/43.6 | 119/10.5 |
| iCaRL (20%) | 924/150 | 299/71.7 | 123/19.1 | 149/16.5 | 228/57.1 | 130/11.0 |
| iCaRL (40%) | - | - | - | - | - | 111/11.9 |
| GEM (1%) | 607/385 | 262/275 | 86.6/53.4 | 117/57.3 | 196/170 | 102/70.2 |
| GEM (5%) | 1085/1012 | 289/377 | 92.3/65.7 | 119/61.9 | 219/224 | 105/81.6 |
| GEM (10%) | 1529/1521 | 379/624 | 94.2/70.1 | 122/71.6 | 295/380 | 104/76.6 |
| GEM (20%) | 2641/2660 | 576/1247 | 132/142 | 124/85.2 | 454/656 | 102/72.2 |
| GEM (40%) | - | - | - | - | - | 106/83.5 |

Table 6: Average Latency (Training Time/IL Time) in seconds for iCaRL on three datasets - Scenario 3 on Smartphone.

| IL Method | Skoda | Ninapro (Per Subject) | EmotionSense |
|---|---|---|---|
| iCaRL (1%) | 4400/9 | 1956/1.28 | 1568/0.5 |
| iCaRL (5%) | 3894/29 | 1974/3 | 1388/1.91 |
| iCaRL (10%) | 3869/72 | 2312/4.5 | 1535/2.6 |
| iCaRL (20%) | 3902/212 | 2008/5.1 | 1517/4.7 |
| iCaRL (40%) | - | - | 1506/8.1 |

**Memory footprint:** We further examine the peak memory usage of iCaRL with its largest budget size of 20-40% on all the datasets to evaluate whether or not it can fit the tight memory budget of Jetson Nano. The peak memory overheads of running the end-to-end IL range from 196 MB for our smallest dataset of EmotionSense to 1,194 MB for our largest dataset of HHAR, when the CPU is used for IL. Then, when we use GPU for running iCaRL, it incurs 1,782-2,127 MB peak memory and requires an additional swap space of 750-3,523 MB. Note that we report the upper bound of the peak memory usage to understand the memory resource requirements of IL methods. Also, the memory overheads can be mitigated by using a smaller batch size and budget size that can fit into resource availability of a target resource-constrained device. Furthermore, we observed that the latency reduction using GPU over CPU is largely consistent between 80-86%, indicating that the swap space has minimal impacts on the speed-up of the IL using GPU compared to using CPU on Jetson Nano. *This result confirms that IL in the mobile* and embedded sensing domain is applicable on resource-constrained devices within a reasonable memory overhead.

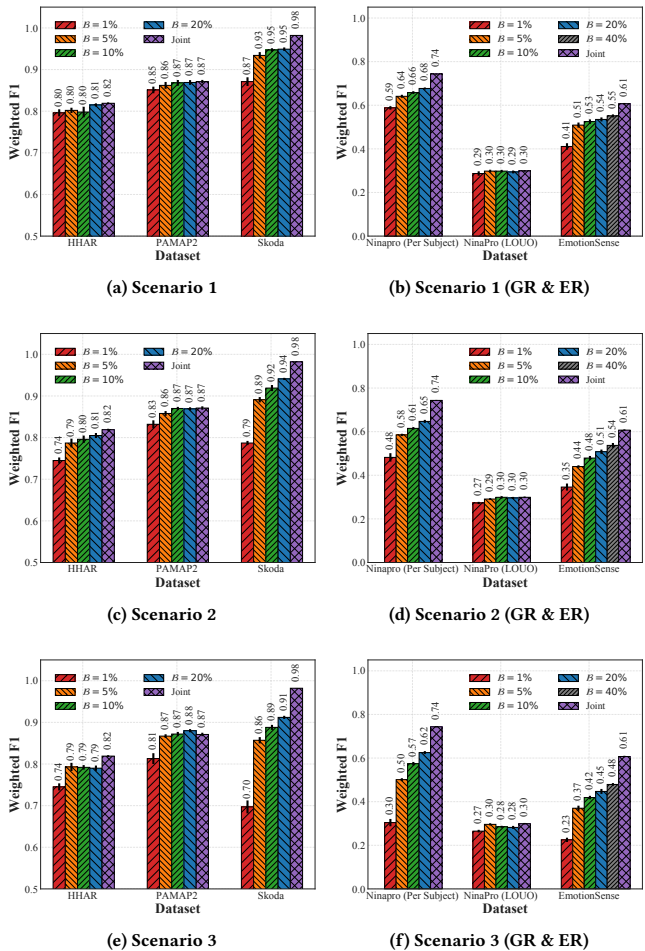

Figure 5: The parameter analysis of the best performing model, iCaRL, in all tasks (HAR, GR, and ER) for all scenarios according to its storage budgets. Reported results are averaged over 10 trials. Standard-error intervals are depicted.

### 5.5 Performance with IL parameters

We study the importance of the storage budget parameter for iCaRL as it is the best performing IL method. Figure 5 shows the weighted F1-score with changing storage budgets of 1%, 5%, 10%, and 20% of total training samples (up to 40% storage budget for the case of ER). In general, more samples are needed to avoid CF as the complexity of the scenario increases. In Scenario 1, only 1% of total samples are needed to achieve similar performance as the joint model. Moreover, in Scenario 2 and 3, the results show that the budget size of 5% is enough to achieve the high performance which is quite close to that of the joint model, although the difficulty of the task increases compared to Scenario 1. In contrast, 10% of samples are required to achieve near joint model's performance (i.e., upper bound performance) in the most challenging setup (Scenario 3).

Note that the performances of iCaRL with the budget size of 5% are often very close to those of iCaRL with budget sizes of 10%, 20%, and 40%. This result indicates that iCaRL enables us to achieve close to the performance of the joint model without requiring excessive storage (less than 30 MB in all datasets in our experiment when a budget size is 5%). Specifically, the required storage of iCaRL with 5% budget size for each dataset (HHAR, PAMAP2, Skoda, Ninapro (Per Subject), Ninapro (LOUO), and EmotionSense corresponds to 28.84, 12.34, 4.19, 0.51, 3.19, and 0.41 MB, respectively. *This is an interesting finding, making iCaRL a good candidate to perform IL on many embedded devices and smartphones with reasonable storage as only a few samples are required to be stored.*

## 6 DISCUSSION

We discuss the potential guidelines ($\mathcal{G}$) for researchers and practitioners in the mobile and embedded systems community based on our findings of this work. The readers should take our results and guidelines with a pinch of salt as we did not compare all the existing IL methods due to reasons mentioned earlier (Section 3) and these findings are based on a few prominent IL methods we analyzed in our study.

- ($\mathcal{G}_1$): If storage is not an issue on the device, one can choose to use the iCaRL method since it performs best across all datasets in different sensing applications. As many modern computing platforms including smartphones and embedded devices have large storage capacity, the issue of storing a proportion of training samples can be minor. iCaRL is also not very computationally expensive on the modern embedded devices and the smartphone. Also, the process can be sped up by using GPUs, although it incurs higher peak memory than CPUs.
- ($\mathcal{G}_2$): GEM, although being a replay with exemplars-based method like iCaRL, should not be preferred over iCaRL as its performance remains inferior to those of iCaRL. Also, GEM is computationally expensive as well as requires more storage than iCaRL.
- ($\mathcal{G}_3$): In a severely resource-constrained environment, EWC and Online EWC can be a reasonable alternative to iCaRL since these methods require less additional storage. Although EWC is a computationally expensive method, the computational cost can be manageable as the IL process is only performed once per task. One can reduce the number of samples used to compute fisher matrices, which account for the majority of the IL time.
- ($\mathcal{G}_4$): LwF and SI should be avoided as they offer minimal protection against CF on mobile sensing applications.
- ($\mathcal{G}_5$): Suppose the available resources such as storage are constrained on the device. In that case, we suggest using iCaRL with a budget size of 1%–5% of training samples as using a higher budget size does not always provide enough benefits if the training dataset size is large (HHAR PAMAP2, and NinaPro (LOUO)). On the other hand, for datasets having smaller training sizes such as Ninapro (Per Subject) and EmotionSense datasets, having a higher budget of 20%–40% helps to a large extent.

# 7 CONCLUSIONS AND FUTURE WORK

In this paper, we studied the CF problem using six prominent IL methods based on three representative sensing applications (i.e., HAR, GR, and ER) in three continual learning scenarios with varying complexities. With our end-to-end IL framework implemented on Nvidia Jetson Nano and a smartphone (OnePlus 7 Pro), we conducted extensive experiments to investigate IL methods' performance, generalizability, and trade-offs of storage, computational costs, and memory footprints. We first identified that CF occurs in mobile and embedded sensing applications when IL methods are not used. We also found that while most IL methods solve the CF in simple scenarios, only iCaRL among the compared methods can successfully alleviate CF issues in more challenging scenarios across the employed datasets. Furthermore, we demonstrated that the IL approaches incur minor to modest storage, peak memory usage, and latency overheads (a minute per task in general), thereby saving a considerable amount of computational resources on-device compared to a case when training is done from scratch whenever a new class/task is added to the system. Finally, based on those findings, we discuss potential guidelines for practitioners and researchers interested in applying IL to edge platforms.

As future work, we believe that it would be worthwhile to further investigate continual learning on more severely resource-constrained devices such as microcontrollers as they have smaller storage, limited memory, and low computational power to apply IL methods. Moreover, we want to study how model compression techniques such as quantization affect IL methods' performance. Similarly, combining binary neural networks with IL methods can be interesting future work. The other key point our study highlighted is that the major bottleneck comes from the training during the IL process. In this context, techniques such as Mixed Precision Training (MPT) [9] and quantization using only 16 or 8-bit floating-point representation [51] for weights might help improve the training efficiency in terms of its computational costs, memory footprints and latency.

## ACKNOWLEDGMENTS

This work is supported by a Google Faculty Award 2019 and by Nokia Bell Labs through their donation for the Centre of Mobile, Wearable Systems and Augmented Intelligence to the University of Cambridge. The authors declare that they have no conflict of interest with respect to the publication of this work.

## REFERENCES

[1] Rahaf Aljundi, Punarjay Chakravarty, and Tinne Tuytelaars. 2017. Expert Gate: Lifelong Learning With a Network of Experts. 3366–3375. http://openaccess.thecvf.com/content_cvpr_2017/html/Aljundi_Expert_Gate_Lifelong_CVPR_2017_paper.html

[2] Manfredo Atzori, Arjan Gijsberts, Claudio Castellini, Barbara Caputo, Anne-Gabrielle Mittaz Hager, Simone Elsig, Giorgio Giatsidis, Franco Bassetto, and Henning Müller. 2014. Electromyography data for non-invasive naturally-controlled robotic hand prostheses. *Scientific Data* 1 (Dec. 2014), 140053. https://doi.org/10.1038/sdata.2014.53

[3] Vincent Becker, Pietro Oldrati, Liliana Barrios, and Gábor Sörös. 2018. Touchsense: Classifying Finger Touches and Measuring Their Force with an Electromyography Armband. In *Proceedings of the 2018 ACM International Symposium on Wearable Computers (ISWC '18)*. ACM, New York, NY, USA, 1–8. https://doi.org/10.1145/3267242.3267250 event-place: Singapore, Singapore.

[4] Sourav Bhattacharya and Nicholas D Lane. 2016. Sparsification and separation of deep learning layers for constrained resource inference on wearables. In *Proceedings of the 14th ACM Conference on Embedded Network Sensor Systems CD-ROM*. 176–189.

[5] Andreas Bulling, Ulf Blanke, and Bernt Schiele. 2014. A Tutorial on Human Activity Recognition Using Body-worn Inertial Sensors. *ACM Comput. Surv.* 46, 3 (Jan. 2014), 33:1–33:33. https://doi.org/10.1145/2499621

[6] Arslan Chaudhry, Puneet K. Dokania, Thalaiyasingam Ajanthan, and Philip H. S. Torr. 2018. Riemannian Walk for Incremental Learning: Understanding Forgetting and Intransigence. 532–547. http://openaccess.thecvf.com/content_ECCV_2018/html/Arslan_Chaudhry__Riemannian_Walk_ECCV_2018_paper.html

[7] Jagmohan Chauhan, Young D. Kwon, Pan Hui, and Cecilia Mascolo. 2020. ContAuth: Continual Learning Framework for Behavioral-based User Authentication. *Proceedings of the ACM on Interactive, Mobile, Wearable and Ubiquitous Technologies* 4, 4 (Dec. 2020), 122:1–122:23. https://doi.org/10.1145/3432203

[8] Guoguo Chen, Carolina Parada, and Georg Heigold. 2014. Small-footprint keyword spotting using deep neural networks. In *2014 IEEE International Conference on Acoustics, Speech and Signal Processing (ICASSP)*. IEEE, 4087–4091.

[9] Dipankar Das, Naveen Mellempudi, Dheevatsa Mudigere, Dhiraj Kalamkar, Sasikanth Avancha, Kunal Banerjee, Srinivas Sridharan, Karthik Vaidyanathan, Bharat Kaul, Evangelos Georganas, et al. 2018. Mixed precision training of convolutional neural networks using integer operations. *arXiv preprint arXiv:1802.00930* (2018).

[10] Junjun Fan, Xiangmin Fan, Feng Tian, Yang Li, Zitao Liu, Wei Sun, and Hongan Wang. 2018. What is That in Your Hand?: Recognizing Grasped Objects via Forearm Electromyography Sensing. *Proc. ACM Interact. Mob. Wearable Ubiquitous Technol.* 2, 4 (Dec. 2018), 161:1–161:24. https://doi.org/10.1145/3287039

[11] Petko Georgiev, Sourav Bhattacharya, Nicholas D. Lane, and Cecilia Mascolo. 2017. Low-resource Multi-task Audio Sensing for Mobile and Embedded Devices via Shared Deep Neural Network Representations. *Proceedings of the ACM on Interactive, Mobile, Wearable and Ubiquitous Technologies* 1, 3 (Sept. 2017), 50:1–50:19. https://doi.org/10.1145/3131895

[12] Petko Georgiev, Nicholas D Lane, Kiran K Rachuri, and Cecilia Mascolo. 2014. Dsp. ear: Leveraging co-processor support for continuous audio sensing on smartphones. In *Proceedings of the 12th ACM Conference on Embedded Network Sensor Systems*. 295–309.

[13] Yu Guan and Thomas Plötz. 2017. Ensembles of Deep LSTM Learners for Activity Recognition Using Wearables. *Proc. ACM Interact. Mob. Wearable Ubiquitous Technol.* 1, 2 (June 2017), 11:1–11:28. https://doi.org/10.1145/3090076

[14] Nils Y. Hammerla, Shane Halloran, and Thomas Plötz. 2016. Deep, Convolutional, and Recurrent Models for Human Activity Recognition Using Wearables. In *Proceedings of the Twenty-Fifth International Joint Conference on Artificial Intelligence (IJCAI'16)*. AAAI Press, 1533–1540. http://dl.acm.org/citation.cfm?id=3060832.3060835 event-place: New York, New York, USA.

[15] Harish Haresamudram, David V. Anderson, and Thomas Plötz. 2019. On the Role of Features in Human Activity Recognition. In *Proceedings of the 23rd International Symposium on Wearable Computers (ISWC '19)*. ACM, New York, NY, USA, 78–88. https://doi.org/10.1145/3341163.3347727 event-place: London, United Kingdom.

[16] Geoffrey E. Hinton, Nitish Srivastava, Alex Krizhevsky, Ilya Sutskever, and Ruslan R. Salakhutdinov. 2012. Improving neural networks by preventing co-adaptation of feature detectors. *arXiv:1207.0580 [cs]* (July 2012). http://arxiv.org/abs/1207.0580 arXiv: 1207.0580.

[17] Sepp Hochreiter and Jürgen Schmidhuber. 1997. Long Short-Term Memory. *Neural Computation* 9, 8 (Nov. 1997), 1735–1780. https://doi.org/10.1162/neco.1997.9.8.1735

[18] Saurav Jha, Martin Schiemer, Franco Zambonelli, and Juan Ye. 2021. Continual learning in sensor-based human activity recognition: An empirical benchmark analysis. *Information Sciences* 575 (Oct. 2021), 1–21. https://doi.org/10.1016/j.ins.2021.04.062

[19] Yifei Jiang, Xin Pan, Kun Li, Qin Lv, Robert P. Dick, Michael Hannigan, and Li Shang. 2012. ARIEL: Automatic Wi-fi Based Room Fingerprinting for Indoor Localization. In *Proceedings of the 2012 ACM Conference on Ubiquitous Computing (UbiComp '12)*. ACM, New York, NY, USA, 441–450. https://doi.org/10.1145/2370216.2370282 event-place: Pittsburgh, Pennsylvania.

[20] Ronald Kemker, Marc McClure, Angelina Abitino, Tyler L Hayes, and Christopher Kanan. 2018. Measuring catastrophic forgetting in neural networks. In *Thirty-second AAAI conference on artificial intelligence*.

[21] Gary King and Langche Zeng. 2001. Logistic Regression in Rare Events Data. *Political Analysis* 9, 2 (2001), 137–163. https://doi.org/10.1093/oxfordjournals.pan.a004868

[22] James Kirkpatrick, Razvan Pascanu, Neil Rabinowitz, Joel Veness, Guillaume Desjardins, Andrei A. Rusu, Kieran Milan, John Quan, Tiago Ramalho, Agnieszka Grabska-Barwinska, Demis Hassabis, Claudia Clopath, Dharshan Kumaran, and Raia Hadsell. 2017. Overcoming catastrophic forgetting in neural networks. *Proceedings of the National Academy of Sciences* 114, 13 (March 2017), 3521–3526. https://doi.org/10.1073/pnas.1611835114

[23] Young D. Kwon, Jagmohan Chauhan, and Cecilia Mascolo. 2021. FastICARL: Fast Incremental Classifier and Representation Learning with Efficient Budget Allocation in Audio Sensing Applications. In *Proc. Interspeech 2021*. 356–360. https://doi.org/10.21437/Interspeech.2021-1091

[24] Young D. Kwon, Kirill A. Shatilov, Lik-Hang Lee, Serkan Kumyol, Kit-Yung Lam, Yui-Pan Yau, and Pan Hui. 2020. MyoKey: Surface Electromyography and Inertial Motion Sensing-based Text Entry in AR. In *2020 IEEE International Conference on Pervasive Computing and Communications Workshops (PerCom Workshops)*. 1–4. https://doi.org/10.1109/PerComWorkshops48775.2020.9156084

[25] Nicholas D. Lane, Petko Georgiev, and Lorena Qendro. 2015. DeepEar: Robust Smartphone Audio Sensing in Unconstrained Acoustic Environments Using Deep Learning. In *Proceedings of the 2015 ACM International Joint Conference on Pervasive and Ubiquitous Computing (UbiComp '15)*. ACM, New York, NY, USA, 283–294. https://doi.org/10.1145/2750858.2804262 event-place: Osaka, Japan.

[26] Sang-Woo Lee, Jin-Hwa Kim, Jaehyun Jun, Jung-Woo Ha, and Byoung-Tak Zhang. 2017. Overcoming Catastrophic Forgetting by Incremental Moment Matching. In *Advances in Neural Information Processing Systems 30*, I. Guyon, U. V. Luxburg, S. Bengio, H. Wallach, R. Fergus, S. Vishwanathan, and R. Garnett (Eds.). Curran Associates, Inc., 4652–4662. http://papers.nips.cc/paper/7051-overcoming-catastrophic-forgetting-by-incremental-moment-matching.pdf

[27] G. Li, A. E. Schultz, and T. A. Kuiken. 2010. Quantifying Pattern Recognition—Based Myoelectric Control of Multifunctional Transradial Prostheses. *IEEE Transactions on Neural Systems and Rehabilitation Engineering* 18, 2 (April 2010), 185–192. https://doi.org/10.1109/TNSRE.2009.2039619

[28] Z. Li and D. Hoiem. 2018. Learning without Forgetting. *IEEE Transactions on Pattern Analysis and Machine Intelligence* 40, 12 (Dec. 2018), 2935–2947. https://doi.org/10.1109/TPAMI.2017.2773081

[29] David Lopez-Paz and Marc\textquotesingle Aurelio Ranzato. 2017. Gradient Episodic Memory for Continual Learning. In *Advances in Neural Information Processing Systems 30*, I. Guyon, U. V. Luxburg, S. Bengio, H. Wallach, R. Fergus, S. Vishwanathan, and R. Garnett (Eds.). Curran Associates, Inc., 6467–6476. http://papers.nips.cc/paper/7225-gradient-episodic-memory-for-continual-learning.pdf

[30] Hong Lu, Denise Frauendorfer, Mashfiqui Rabbi, Marianne Schmid Mast, Gokul T. Chittaranjan, Andrew T. Campbell, Daniel Gatica-Perez, and Tanzeem Choudhury. 2012. StressSense: Detecting Stress in Unconstrained Acoustic Environments Using Smartphones. In *Proceedings of the 2012 ACM Conference on Ubiquitous Computing (UbiComp '12)*. ACM, New York, NY, USA, 351–360. https://doi.org/10.1145/2370216.2370270 event-place: Pittsburgh, Pennsylvania.

[31] James L. McClelland, Bruce L. McNaughton, and Randall C. O'Reilly. 1995. Why there are complementary learning systems in the hippocampus and neocortex: Insights from the successes and failures of connectionist models of learning and memory. *Psychological Review* 102, 3 (1995), 419–457. https://doi.org/10.1037/0033-295X.102.3.419

[32] Michael McCloskey and Neal J. Cohen. 1989. Catastrophic Interference in Connectionist Networks: The Sequential Learning Problem. In *Psychology of Learning and Motivation*, Gordon H. Bower (Ed.). Vol. 24. Academic Press, 109–165. https://doi.org/10.1016/S0079-7421(08)60536-8

[33] Henry Friday Nweke, Ying Wah Teh, Mohammed Ali Al-garadi, and Uzoma Rita Alo. 2018. Deep learning algorithms for human activity recognition using mobile and wearable sensor networks: State of the art and research challenges. *Expert Systems with Applications* 105 (Sept. 2018), 233–261. https://doi.org/10.1016/j.eswa.2018.03.056

[34] Francisco Javier Ordóñez and Daniel Roggen. 2016. Deep Convolutional and LSTM Recurrent Neural Networks for Multimodal Wearable Activity Recognition. *Sensors* 16, 1 (Jan. 2016), 115. https://doi.org/10.3390/s16010115

[35] German I. Parisi, Ronald Kemker, Jose L. Part, Christopher Kanan, and Stefan Wermter. 2019. Continual lifelong learning with neural networks: A review. *Neural Networks* 113 (May 2019), 54–71. https://doi.org/10.1016/j.neunet.2019.01.012

[36] B. Pfülb and A. Gepperth. 2019. A comprehensive, application-oriented study of catastrophic forgetting in DNNs. In *ICLR*.

[37] Angkoon Phinyomark and Erik Scheme. 2018. EMG Pattern Recognition in the Era of Big Data and Deep Learning. *Big Data and Cognitive Computing* 2, 3 (Sept. 2018), 21. https://doi.org/10.3390/bdcc2030021

[38] Hendrik Purwins, Bo Li, Tuomas Virtanen, Jan Schlüter, Shuo-Yiin Chang, and Tara Sainath. 2019. Deep Learning for Audio Signal Processing. *IEEE Journal of Selected Topics in Signal Processing* 13, 2 (May 2019), 206–219. https://doi.org/10.1109/JSTSP.2019.2908700

[39] Kiran K. Rachuri, Mirco Musolesi, Cecilia Mascolo, Peter J. Rentfrow, Chris Longworth, and Andrius Aucinas. 2010. EmotionSense: a mobile phones based adaptive platform for experimental social psychology research. In *Proceedings of the 12th ACM international conference on Ubiquitous computing (UbiComp '10)*. Association for Computing Machinery, Copenhagen, Denmark, 281–290. https://doi.org/10.1145/1864349.1864393

[40] Sylvestre-Alvise Rebuffi, Alexander Kolesnikov, Georg Sperl, and Christoph H Lampert. 2017. icarl: Incremental classifier and representation learning. In *Proceedings of the IEEE conference on Computer Vision and Pattern Recognition*. 2001–2010.

[41] A. Reiss and D. Stricker. 2012. Introducing a New Benchmarked Dataset for Activity Monitoring. In *2012 16th International Symposium on Wearable Computers*. 108–109. https://doi.org/10.1109/ISWC.2012.13

[42] Monika Schak and Alexander Gepperth. [n.d.]. A study on catastrophic forgetting in deep LSTM networks. ([n. d.]), 14.

[43] Jonathan Schwarz, Wojciech Czarnecki, Jelena Luketina, Agnieszka Grabska-Barwinska, Yee Whye Teh, Razvan Pascanu, and Raia Hadsell. 2018. Progress & Compress: A scalable framework for continual learning. In *International Conference on Machine Learning*. 4535–4544.

[44] Jonathan Schwarz, Wojciech Czarnecki, Jelena Luketina, Agnieszka Grabska-Barwinska, Yee Whye Teh, Razvan Pascanu, and Raia Hadsell. 2018. Progress & Compress: A scalable framework for continual learning. In *International Conference on Machine Learning*. 4528–4537. http://proceedings.mlr.press/v80/schwarz18a.html

[45] Sandra Servia-Rodriguez, Cecilia Mascolo, and Young D. Kwon. 2021. Knowing when we do not know: Bayesian continual learning for sensing-based analysis tasks. *arXiv:2106.05872 [cs]* (June 2021).

[46] Kirill A. Shatilov, Dimitris Chatzopoulos, Alex Wong Tat Hang, and Pan Hui. 2019. Using Deep Learning and Mobile Offloading to Control a 3D-printed Prosthetic Hand. *Proc. ACM Interact. Mob. Wearable Ubiquitous Technol.* 3, 3 (Sept. 2019), 102:1–102:19. https://doi.org/10.1145/3351260

[47] M. Smith and T. Barnwell. 1987. A new filter bank theory for time-frequency representation. *IEEE Transactions on Acoustics, Speech, and Signal Processing* 35, 3 (March 1987), 314–327. https://doi.org/10.1109/TASSP.1987.1165139 Conference Name: IEEE Transactions on Acoustics, Speech, and Signal Processing.

[48] Thomas Stiefmeier, Daniel Roggen, Georg Ogris, Paul Lukowicz, and Gerhard Tröster. 2008. Wearable Activity Tracking in Car Manufacturing. *IEEE Pervasive Computing* 7, 2 (April 2008), 42–50. https://doi.org/10.1109/MPRV.2008.40

[49] Allan Stisen, Henrik Blunck, Sourav Bhattacharya, Thor Siiger Prentow, Mikkel Baun Kj\a ergaard, Anind Dey, Tobias Sonne, and Mads Møller Jensen. 2015. Smart Devices Are Different: Assessing and MitigatingMobile Sensing Heterogeneities for Activity Recognition. In *Proceedings of the 13th ACM Conference on Embedded Networked Sensor Systems (SenSys '15)*. ACM, New York, NY, USA, 127–140. https://doi.org/10.1145/2809695.2809718 event-place: Seoul, South Korea.

[50] Gido M. van de Ven and Andreas S. Tolias. 2019. Three scenarios for continual learning. *arXiv:1904.07734 [cs, stat]* (April 2019). http://arxiv.org/abs/1904.07734 arXiv: 1904.07734.

[51] Naigang Wang, Jungwook Choi, Daniel Brand, Chia-Yu Chen, and Kailash Gopalakrishnan. 2018. Training deep neural networks with 8-bit floating point numbers. In *Advances in neural information processing systems*. 7675–7684.

[52] Max Welling. 2009. Herding dynamical weights to learn. In *Proceedings of the 26th Annual International Conference on Machine Learning (ICML '09)*. Association for Computing Machinery, Montreal, Quebec, Canada, 1121–1128. https://doi.org/10.1145/1553374.1553517

[53] Shuochao Yao, Shaohan Hu, Yiran Zhao, Aston Zhang, and Tarek Abdelzaher. 2017. DeepSense: A Unified Deep Learning Framework for Time-Series Mobile Sensing Data Processing. In *Proceedings of the 26th International Conference on World Wide Web (WWW '17)*. International World Wide Web Conferences Steering Committee, Republic and Canton of Geneva, Switzerland, 351–360. https://doi.org/10.1145/3038912.3052577 event-place: Perth, Australia.

[54] Jaehong Yoon, Eunho Yang, Jeongtae Lee, and Sung Ju Hwang. 2018. Lifelong Learning with Dynamically Expandable Networks. (Feb. 2018). https://openreview.net/forum?id=Sk7KsfW0-

[55] Friedemann Zenke, Ben Poole, and Surya Ganguli. 2017. Continual Learning Through Synaptic Intelligence. In *Proceedings of the 34th International Conference on Machine Learning - Volume 70 (ICML'17)*. JMLR.org, 3987–3995. http://dl.acm.org/citation.cfm?id=3305890.3306093 event-place: Sydney, NSW, Australia.

[56] Xiaolong Zhai, Beth Jelfs, Rosa H. M. Chan, and Chung Tin. 2017. Self-Recalibrating Surface EMG Pattern Recognition for Neuroprosthesis Control Based on Convolutional Neural Network. *Frontiers in Neuroscience* 11 (2017). https://doi.org/10.3389/fnins.2017.00379